\documentclass[nonacm]{acmart}

\usepackage[utf8]{inputenc}
\usepackage{times}
\usepackage{amsmath,mathrsfs,mathtools}
\usepackage{graphicx} 
\usepackage{caption}
\usepackage{xspace}
\usepackage{soul}
\usepackage{geometry}
\usepackage{float}
\usepackage{booktabs}
\usepackage{multirow}
\usepackage{algorithm}
\usepackage[noend]{algpseudocode}
\usepackage{listings}
\usepackage{subfig}
\usepackage{cleveref}
\usepackage[footnotes]{trackchanges}
\usepackage{etoolbox}
\usepackage{ifthen}
\usepackage{adjustbox}
\usepackage{makecell}
\usepackage{xspace}
\usepackage{siunitx}

\definecolor[named]{MyPink}{cmyk}{0.0,1.0,0.5,.13}
\definecolor[named]{MyPurple}{cmyk}{0.5,1.,0,0.4}
\definecolor[named]{MyRed}{cmyk}{0.00,0.17,0.18,0.03}
\definecolor[named]{MyDarkRed}{cmyk}{0.00,0.54,0.57,0.40}
\definecolor[named]{MyGreen}{cmyk}{0.08,0.00,0.09,0.09}
\definecolor[named]{MyDarkGreen}{cmyk}{0.27,0.00,0.43,0.40}
\definecolor[named]{MyGrey}{cmyk}{0.00,0.00,0.00,0.40}

\newcommand{\ph}[1]{\texttt{\textcolor{ACMOrange}{#1}}}

\newtoggle{archall}
\togglefalse{archall}

\def\cvprPaperID{****} 
\def\confName{ICCV}
\def\confYear{2021}
\graphicspath{{./figs/}{./figs/result/}}

\newcommand{\net}[1]{\Phi_{\theta}(#1)}

\newcommand{\norm}[1]{\lVert #1 \rVert_2}
\newcommand{\bestb}[1]{$\mb{#1}$}
\newcommand{\bestn}[1]{$\mb{#1}$}

\newcommand{\nR}{\mathbb{R}}

\newcommand{\cB}{\mathcal{B}}

\newcommand{\cI}{\mathcal{I}}

\newcommand{\cL}{\mathcal{L}}

\newcommand{\cP}{\mathcal{P}}

\newcommand{\cR}{\mathcal{R}}

\newcommand{\cT}{\mathcal{T}}
\newcommand{\cU}{\mathcal{U}}

\newcommand{\cY}{\mathcal{Y}}

\newcommand{\figref}[1]{Fig.~\ref{#1}}
\newcommand{\secref}[1]{Section~\ref{#1}}

\newcommand{\tabref}[1]{Table~\ref{#1}}

\def\mb{\mathbf}

\makeatletter
\DeclareRobustCommand\onedot{\futurelet\@let@token\@onedot}
\def\@onedot{\ifx\@let@token.\else.\null\fi\xspace}

\makeatother

\definecolor{darkred}{rgb}{0.8,0,0}
\definecolor{darkgreen}{rgb}{0,0.5,0}
\definecolor{darkblue}{rgb}{0,0,0.7}
\definecolor{darkpurple}{rgb}{0.4,0,0.6}
\definecolor{lightgray}{rgb}{0.92,0.92,0.92}
\definecolor{lightpink}{rgb}{1.00,0.90,0.90}

\definecolor{ellisred}{rgb}{0.87,0.44,0.38}
\definecolor{ellisgreen}{rgb}{0.69,0.90,0.52}
\definecolor{elliscyan}{rgb}{0.29,0.77,0.74}
\definecolor{ellisorange}{rgb}{0.89,0.55,0.28}
\definecolor{ellisblue}{rgb}{0.41,0.61,0.86}

\newcommand{\imT}{\ensuremath{I^T}\xspace}
\newcommand{\imV}{\ensuremath{I^E}\xspace}
\newcommand{\im}{\ensuremath{I}\xspace}
\newcommand{\imdef}{\ensuremath{\im = \imT \cup \imV}\xspace}
\newcommand{\yb}{\ensuremath{\cY_b}\xspace}
\newcommand{\yn}{\ensuremath{\cY_n}\xspace}
\newcommand{\y}{\ensuremath{\cY}\xspace}
\newcommand{\ydef}{\ensuremath{\y = \yb \cup \yn}\xspace}
\newcommand{\xT}{\cB^T}
\newcommand{\uT}{\cU^T}
\newcommand{\xV}{\cB^E}
\newcommand{\uV}{\cU^E}
\newcommand{\xTdef}{\ensuremath{\cB^T = \{(I,y); I \in \imT, y \in \yb \}}\xspace}
\newcommand{\uTdef}{\ensuremath{\cU^T = \{(I,y); I \in \imT, y \in \yn \}}\xspace}

\def\PlainSupervisedLong{Vanilla\xspace}
\def\TripletLossLong{Triplet Loss\xspace}
\def\ClassificationWithTripletLossLong{Classification With Triplet Loss\xspace}

\def\SemiSupervisedLong{Classification With Rotation\xspace}

\def\PlainSupervised{Vanilla\xspace}
\def\TripletLoss{Triplet\xspace}
\def\ClassificationWithTripletLoss{CwT\xspace}
\def\UnsupervisedRotation{RotNet\xspace}
\def\SemiSupervised{CwRot\xspace}
\def\SimSiam{SimSiam\xspace}
\def\SupContrast{SupContrast\xspace}
\def\FixMatch{FixMatch\xspace}

\def\Cif{CIFAR10\xspace}
\def\CIF{CIFAR100\xspace}
\def\In{ImageNet100\xspace}
\def\IN{ImageNet\xspace}
\def\FC100{\text{Few-Shot} \CIF}
\def\tiered{\textit{tiered}\IN}

\newcommand{\vehicle}[1]{\textcolor{MyPurple}{\bf{#1}}}
\newcommand{\animal}[1]{\textcolor{MyPink}{\bf{#1}}}

\def\ncr{novel class retrieval\xspace}
\def\NCR{Novel Class Retrieval\xspace}
\def\B{base\xspace}
\def\N{novel\xspace}
\def\BC{\B classes\xspace}
\def\NC{\N classes\xspace}
\def\RS{random split\xspace}
\def\SS{semantic split\xspace}

\setcopyright{none}

\AtBeginDocument{%
  \providecommand\BibTeX{{%
    \normalfont B\kern-0.5em{\scshape i\kern-0.25em b}\kern-0.8em\TeX}}}

\begin{document}


\begin{abstract}
Supervised representation learning with deep networks 
tends to overfit the training classes 
and the generalization to \NC is a challenging question. 
It is common to evaluate a learned embedding on held-out images of the same training classes. 
In real applications however, data comes from new sources and 
\NC are likely to arise. 
We hypothesize that incorporating unlabelled images of \NC 
in the training set in a semi-supervised fashion 
would be beneficial for the efficient retrieval of novel-class images 
compared to a vanilla supervised representation. 
To verify this hypothesis in a comprehensive way, 
we propose an original evaluation methodology 
that varies the degree of novelty of \NC 
by partitioning the dataset category-wise either randomly, 
or semantically, 
i.e. by minimizing the shared semantics between \B and \N classes. 
This evaluation procedure allows to train a representation 
blindly to any \N-class labels and evaluate the frozen representation on the retrieval of \B or \N classes.
We find that a vanilla supervised representation falls short on the retrieval
of \NC even more so when the semantics gap is higher.
Semi-supervised algorithms allow to partially bridge this performance gap but there is still much room for improvement.

\end{abstract}

\title{
How does the degree of novelty impacts semi-supervised representation learning for \ncr?
}

\newtoggle{cameraready}
\toggletrue{cameraready}
\iftoggle{cameraready}{
\author{Quentin Leroy}
\affiliation{
    \institution{Inria}
    \city{Montpellier}
    \country{France}
}
\email{quentin.leroy@inria.fr}

\author{Olivier Buisson}
\affiliation{%
    \institution{Institut National de l'Audiovisuel}
    \city{Bry-sur-Marne}
    \country{France}
}
\email{obuisson@ina.fr}

\author{Alexis Joly}
\affiliation{%
    \institution{Inria}
    \city{Montpellier}
    \country{France}
}
\email{alexis.joly@inria.fr}
}{
\author{Anonymous}
\affiliation{
    \institution{Anonymous}
    \city{Anonymous}
    \country{Anonymous}
}
\email{Anonymous}
}


\maketitle
\section{Introduction}
Deep neural networks have been the de facto algorithm
for processing and training data
for a variety of tasks
and data modalities whether it is 
image
~\cite{oquab2014,razavian2014a},
text
\cite{vaswani2017,devlin2019},
speech
\cite{hannun2014,zhang2020}
, 
video 
\cite{aytar2016,sun2019a,alayrac2020,arnab2021}.
In this study we focus on image data and retrieval tasks.
The major impediment
is the sheer amount of annotated images necessary 
to achieve this feat
~\cite{russakovsky2015a}.
Several lines of research 
have been tackling this issue 
and have brought new ideas 
helping to reduce the amount of annotation necessary:
active learning refers to algorithms that select most effectively the unlabeled samples for an annotator to annotate
\cite{settles2010},
few-shot learning refers to algorithms that adapt most rapidly to new classes with a few annotated support samples
\cite{vinyals2017,wang2019},
self-supervised learning refers to algorithms that learn from the samples themselves without any annotation
\cite{doersch2015,doersch2017,jing2019},
semi-supervised learning refers to algorithms that effectively leverage unlabeled data while also training with supervised data
\cite{chapelle2006,laine2016,tarvainen2017,phama2021}.


Deep networks provide a strong baseline 
for image representation in image retrieval applications
~\cite{babenko2014}~\cite{babenko2015},
whether it is visual instance retrieval 
or more broadly visual category retrieval.
Image retrieval, a long tradition in the computer vision community, 
is the task of retrieving those images in the search corpus 
that depict the same visual \textit{instance} as the query.
Visual object category retrieval 
evaluates more broadly the retrieving of those images 
containing the same visual \textit{object} as the query, 
allowing for a lot more appearance variation 
between the query and the correct images.

In this paper, we are interested in the retrieval 
of an object category 
unseen beforehand in the training corpus used to train the deep embedding.
We term this challenging task 
\emph{novel class retrieval};
it aims to reflect the setting of real-world applications 
where incoming image data contains visual content 
novel compared to curated image corpora used for training deep networks.
This setting exists in various applications.
Take for example a corpus of digitized material 
of a cultural institution containing a large variety of 
historical photographs, drawings, paintings and manuscripts of various eras.
Choosing a generic embedding for these original materials 
for efficient retrieval is a challenging question.
These novel materials can be queried by users in unanticipated ways.
Understanding these new types of material 
without additional human annotation is a challenging problem.

We focus on the question: 
What learning algorithm allows for the most efficient retrieval of \NC?
We devise an original methodology 
for comparing deep embeddings towards the task of \ncr;
We partition an image corpus category-wise, 
keeping a set of \BC for training the deep embedding 
and leaving aside a set of \NC
for evaluating the embedding.
We get a strict disjoint label space 
between the training classes and the evaluation classes.
The base and novel sets are further split 
into a training and a test set.
More importantly, 
in each instance of our experiments 
we carry out two kinds of category splits:
a \emph{\RS} 
partitioning the labeled space at random
and a more carefully crafted 
\emph{\SS} 
that minimizes the information overlap between base and unseen categories to the extent possible
(details in~\cref{sec:method}).

Our contributions are summarised as follows: 
1) We devise an evaluation methodology to evaluate a deep embedding on the task of \ncr, 
2) we simulate new visual content by hiding supervision from the deep network and we are able to vary the degree of novelty of \NC and 
3) we compare several deep learning algorithms from different paradigms (supervised, self-supervised, semi-supervised) on the task of \ncr based on three different datasets (\Cif, \CIF, In).

The remainder of the paper is organized as follows: 
the next section reviews important lines of research related to our evaluation methodology.
In~\cref{sec:method} we detail how we manage to construct a benchmark relevant to \ncr by masking whole categories from the training process and varying the degree of novelty.
Next we present the algorithms we implemented and the different datasets and category splits used in the experiments.
Finally we present our experiment results and analysis.

\section{Related Work} \label{sec:related}
\paragraph{\NCR}
We aim to benchmark training algorithms for the task of retrieving samples of classes unseen during training.
It is similar to 
Novel class discovery~\cite{han2021a,zhong2021}
where the goal is to cluster the whole set of images of \NC:
instead of clustering we evaluate with retrieval, we do not need to know the number of novel classes or to estimate it.
The goal in 
Few-shot learning~\cite{vinyals2017,snell2017,oreshkin2019,wang2019}
is to adapt most rapidly to \NC with a small set of support supervised
samples of those \NC: in contrast we evaluate class retrieval, that is returning the most images of the class of a given single image query.
Metric learning~\cite{kulis2013,krause2013,bell2015,schroff2015,song2016,musgrave2020}
benchmarks usually report performance in retrieval tasks of \NC;
datasets for those benchmarks~\cite{krause2013,song2016,wah2011} also have class-disjoint training and testing sets
but do not keep a split along the image axis as we do (See~\figref{fig:datasplits});
in the testing phase we report retrieval performance on test images
of \BC and \NC among both \B and \N-class images while metric learning does not keep test samples of \BC during testing
(which prevents from evaluating the performance decrease compared to \BC).

\paragraph{Semi-Supervised Setting}
It is often pointed out that
labels are expensive to obtain and 
that is why training data is abundant in unlabeled samples
and is lacking labeled samples.
Semi-Supervised learning algorithms~\cite{chapelle2006}
are often proclaimed to work efficiently in that setting~\cite{laine2016,tarvainen2017,li2021b,phama2021},
some methods showing impressive performance
with a quantity of labels as low as one per class for \Cif~\cite{sohn2020a},
or 1\% of labeled data for \IN~\cite{zhai2019}.
We investigate an overlooked setting where
the labeled pool and the unlabeled pool are class-disjoint;
meaning that the unlabeled data are known to come from different classes
that those seen during training.
The authors in~\cite{oliver2018} noted that this should 
be part of an effective benchmark for evaluating semi-supervised learning algorithms.
They did a set of experiments on \Cif where
they vary the amount of classes
present both in labeled and unlabeled data and
showed that performance can decrease in some cases.
We take this idea further in the following ways: 
1) we keep the labeled and unlabeled pools completely class-disjoint,
2) we experiment on \Cif, \CIF and \In and
3) we vary the degree of novelty between labeled and unlabeled pools by
carrying a \RS and a \SS of the label space.

\paragraph{Partitioning the label space semantically}
We experiment \ncr on classification datasets: 
\Cif, \CIF~\cite{krizhevsky2009}
and a subset of \IN~\cite{russakovsky2015a}.
For \CIF we can use the super-classes which group together classes that are visually alike:
we borrow the splits from~\cite{oreshkin2019}
to devised the semantic split.
For \IN we can use the WordNet hierarchy~\cite{fellbaum1998}
; we chose a subset of the high-level categories used for \tiered~\cite{ren2018}
(details in~\secref{sec:results-imagenet100}).

\section{Proposed Evaluation methodology: Varying the degree of novelty}\label{sec:method}
\subsection{Random and Semantic Class-disjoint Data Splits}

\begin{figure}[ht]
    \centering
    \includegraphics[width=\linewidth]{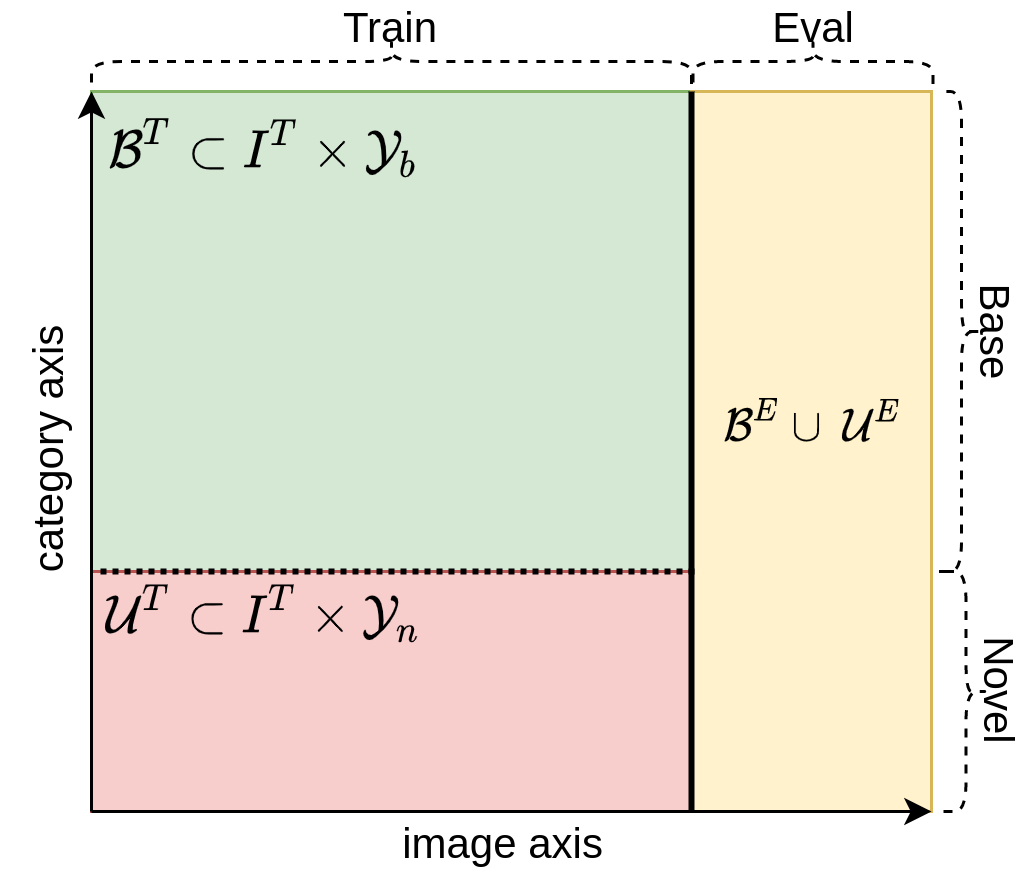}
    \caption{We partition the label space into $\ydef$, 
    holding out a set of novel classes $\yn$ for evaluation. 
    Contrary to metric learning benchmarks 
    we also keep a train/test partition along the image axis $\imdef$, 
    allowing for a comparative performance evaluation 
    on base $\yb$ and novel $\yn$ classes on the test images $\imV$. 
    We train a various set of algorithms 
    with supervision only on base classes $\xT$, 
    and optionally adding images of novel classes 
    of the train partition without supervision $\uT$.}
    \label{fig:datasplits}
\end{figure}

\def\datasplitscaption{
    Illustration of the two training schemes: 
    We train algorithms with supervision on \BC 
    (top row\iftoggle{splitstable}{}{
    ~\protect\subref{fig:datasplits-random-ps},
    ~\protect\subref{fig:datasplits-semantic-ps}
}) 
    and algorithms that incorporate images of the \NC without supervision 
    (bottom row\iftoggle{splitstable}{}{
    ~\protect\subref{fig:datasplits-random-s4ls},
    ~\protect\subref{fig:datasplits-semantic-s4ls}
}). 
    When the label space is partitioned randomly 
    (left column\iftoggle{splitstable}{}{
    ~\protect\subref{fig:datasplits-random-ps},
    ~\protect\subref{fig:datasplits-random-s4ls}, 
} \vehicle{airplane}/\animal{cat} vs. \vehicle{automobile}/\animal{dog}) 
    the semantic gap between \B and \N classes 
    is narrower than when the label space is partitioned semantically 
    (right column\iftoggle{splitstable}{}{
    ~\protect\subref{fig:datasplits-semantic-ps},
    ~\protect\subref{fig:datasplits-semantic-s4ls}, 
} \vehicle{airplane}/\vehicle{automobile} vs. \animal{cat}/\animal{dog}).}

\def\splitwidth{0.40\linewidth}

\newtoggle{splitstable}
\toggletrue{splitstable}

\newcommand{\splitsubfloat}[3]{
    \iftoggle{splitstable}{
        \raisebox{-.5\height}{\includegraphics[width=\splitwidth]{images/datasplits/final/#2/#3}}
        }{
\subfloat[#1]{
        \includegraphics[width=\splitwidth]{images/datasplits/final/#2/#3}
        \label{fig:datasplits-#2-#3}
}
}
}

\newcommand{\firstcol}[1]{\textbf{#1}}

\iftoggle{splitstable}{
\begin{table*}
    \begin{tabular}{ccc}
        \toprule\\
        & \textbf{Random} & \textbf{Semantic} \\
        \firstcol{Supervised} &      \splitsubfloat{}{random}{s} & \splitsubfloat{}{semantic}{s}\\
        \firstcol{Semi-Supervised} & \splitsubfloat{}{random}{ss} & \splitsubfloat{}{semantic}{ss}\\
        \bottomrule
    \end{tabular}
    \caption{\datasplitscaption}
    \label{tab:datasplits}
\end{table*}
}{
\begin{figure*}[ht]
    \centering
    \splitsubfloat{Vanilla Supervised representation learning\\Random split}{random}{s}
    \splitsubfloat{Vanilla Supervised representation learning\\Semantic split}{semantic}{s}
    \qquad
    \splitsubfloat{Semi Supervised representation learning\\Random split}{random}{ss}
    \splitsubfloat{Semi Supervised representation learning\\Semantic split}{semantic}{ss}
    \caption{\datasplitscaption}
    \label{fig:datasplits2}
\end{figure*}
}

We are given an image set $\imdef$ partitioned into training $\imT$ and testing $\imV$ sets (\figref{fig:datasplits}).
We consider that the label space $\y$ is partitioned into two parts:
a set $\yb$ of \BC and a set $\yn$ of \NC.
The training set can be partitioned into $\imT = \xT \cup \uT$
where $\xTdef$ comprises annotated images of base classes
and $\uTdef$ comprises unannotated images of novel classes.
The test set can be partitioned into $\imV = \xV \cup \uV$.

An image $I$ is encoded as $x=\net{I} \in \nR^d$,
where $\net{\cdot}$ is a neural network 
whose parameters $\theta$ are trained on $\xT$ 
and optionally on $\uT$.
The labels of $\uT$ are never used for training the network.

In each dataset, two kinds of experiments are carried out involving either a \RS or a \SS. 
The \RS divides randomly the set of classes.
The \SS divides the set of classes in a way that minimizes the information overlap between the \B and the \N classes.
\tabref{tab:datasplits} illustrates how the \SS increases the degree of novelty compared to the \RS with classes of \Cif.
We evaluate the trained network via image retrieval by running a similarity search
among the whole testing set $\imT$ and report R-Precision 
for queries from \BC $\xV$ and from \NC $\uV$ separately
(see~\secref{sec:results} for the results).

\subsection{Evaluation via Image Retrieval: R-Precision} \label{sec:rprecision}

We want to evaluate 
how well a visual representation 
transfer to unseen classes 
without any additional training, 
contrary to other evaluation methods
that allow for additional training on the novel-class labels
(linear probes~\cite{doersch2015}, 
few-shot tasks~\cite{gidaris2019}, 
knn-classifier~\cite{wu2018}).
Evaluation via clustering requires embeddings to be clustered 
and is dependent on the clustering algorithm used~\cite{musgrave2020}.
We have chosen an evaluation metric 
that does not require any additional metric 
nor a choice of a clustering algorithm.

\paragraph{R-Precision}

Visual object category retrieval aims to 
retrieve examples of the same category 
than a given query from a search set ($\imV$ in our case). 
A function 
$R: I \in \imV \mapsto I_R \in \cP(\imV)$
maps a query image to a set of similar images.
The similarity between two images 
is evaluated by a dot product
in the $\ell_2$-norm normalized feature space 
(also known as cosine similarity):
\begin{equation}
    s(I_i, I_j) = \frac
    {\net{I_i}}
    {\norm{\net{I_i}}} 
    \cdot 
    \frac{\net{I_j}}
    {\norm{\net{I_j}}}
\end{equation}
For an image $I$, $R$ returns the list of most similar images
ranked by the function $s$ in decreasing order.
Let us denote the samples of class $c$ 
by $\cI^c=\{(I_j,y_j), I_j \in \imV, y_j=c\}$,
and the size of the class by 
$N^c=\vert\cI^c\vert$.
Image retrieval performance 
is usually evaluated by 
Recall@$k$~\cite{jegou2011} (with $k=1,5$), 
in the metric learning literature~\cite{song2016}.
We evaluate instead with Precision@$N^c$ where $N^c$ is the total size of correct hits.
The Precision is the proportion of correct hits among the $N^c$ retrieved items.
We set the cut-off-rank to be the size of the class query, 
so that the precision is equal to the recall.

\section{Algorithms Studied} \label{sec:algorithms}
\newtoggle{archone}
\togglefalse{archone}
\newtoggle{archtable}
\toggletrue{archtable}

\renewcommand\theadalign{bc}
\renewcommand\theadfont{\bfseries}
\renewcommand\theadgape{\Gape[4pt]}
\renewcommand\cellgape{\Gape[4pt]}

\iftoggle{archone}{ \def\algofigenv{figure} } { \def\algofigenv{figure*} }

\newcommand{\archwidth}{0.40\linewidth}

\newcommand{\archsubfloat}[2]{
    \iftoggle{archtable}{
        \makecell{ \raisebox{-.5\height}{ \includegraphics[width=\archwidth]{images/arch/#2} } \\ \textbf{#1} }
        }
        { \subfloat[#1]{ \includegraphics[width=\archwidth]{images/arch/#2} } }
}

\def\archcaption{
    Supervised algorithms train a network with images of \BC only (classes \textcolor{MyDarkGreen}{$c_1,c_2$}).
    Unsupervised algorithms train a network with images of \BC without supervision (classes \textcolor{MyDarkRed}{$c_1,c_2$}).
    Semi-Supervised algorithms train a network with images of \BC with supervision, and images of \NC without supervision (classes \textcolor{MyDarkGreen}{$c_1,c_2$} and \textcolor{MyDarkRed}{$c_3,c_4$}).
        At testing time all the \textcolor{MyGrey}{heads} are discarded, we take as a visual representation the output of the last average pooling layer of the backbone $\net{\cdot}$.
}

\iftoggle{archtable}{ 
\begin{table*}[ht]
    \centering
    \begin{tabular}{ccc}
        \toprule \\
        \firstcol{Supervised} & \archsubfloat{\PlainSupervised}{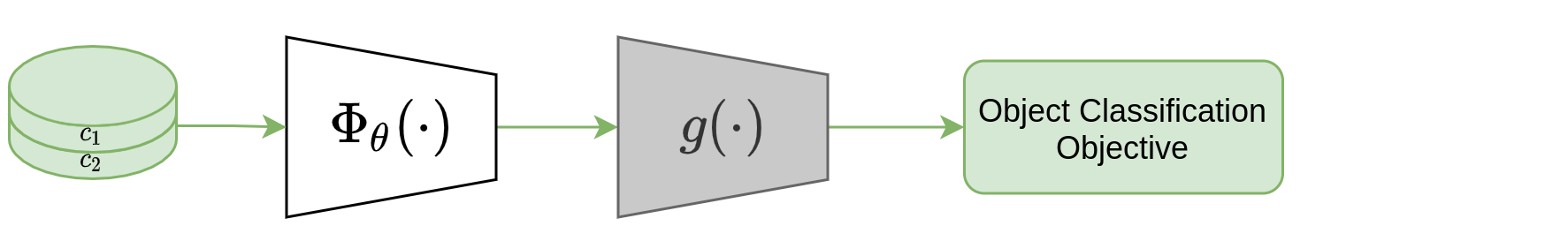} & \archsubfloat{\TripletLoss}{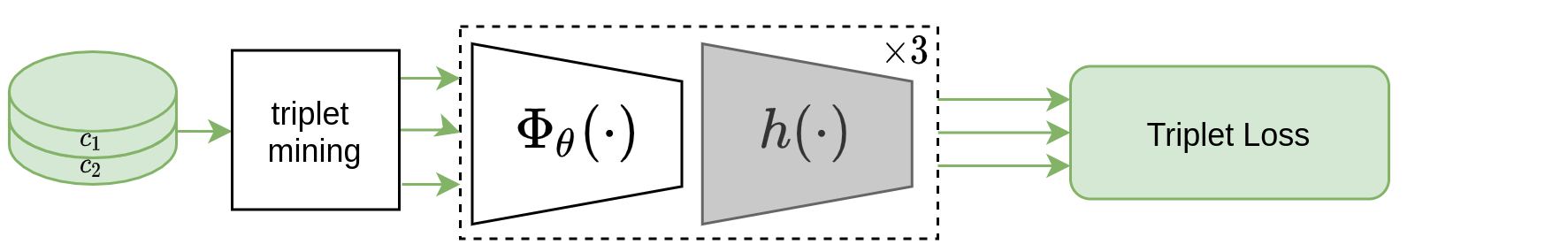} \\
        & \archsubfloat{\ClassificationWithTripletLoss}{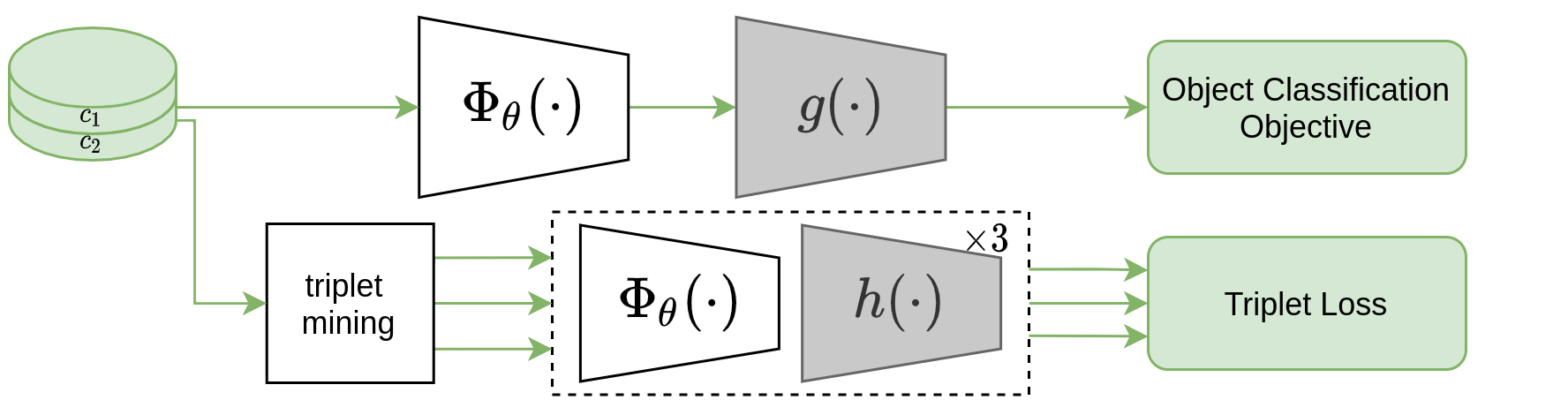} & \archsubfloat{\SupContrast}{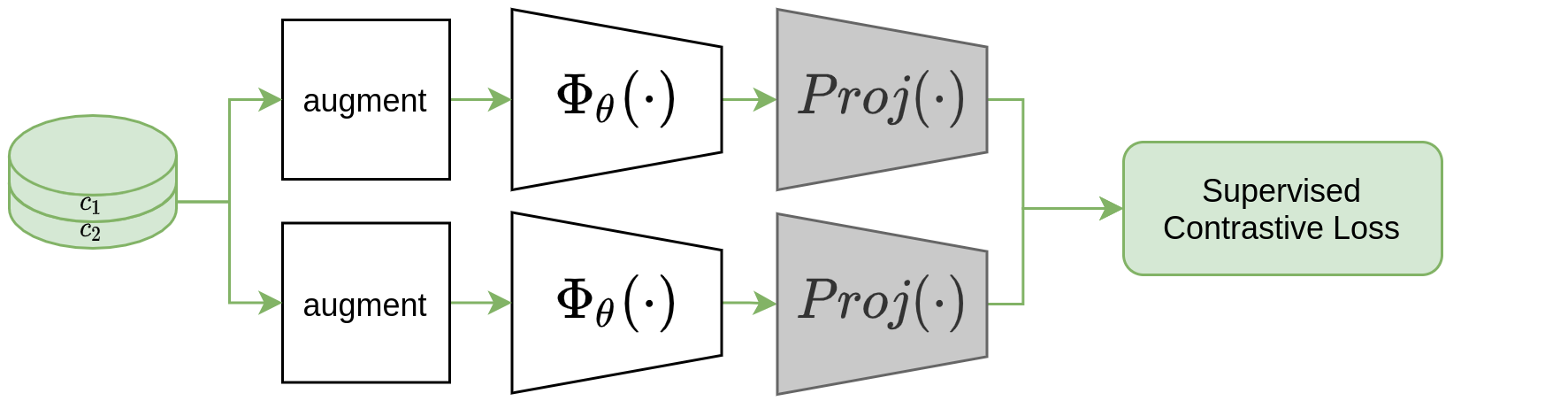} \\
                   \midrule
        \firstcol{Unsupervised} & \archsubfloat{\UnsupervisedRotation}{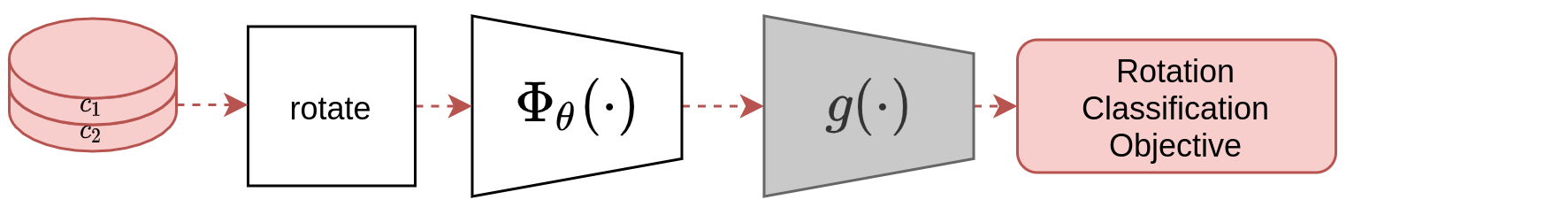} & \archsubfloat{\SimSiam}{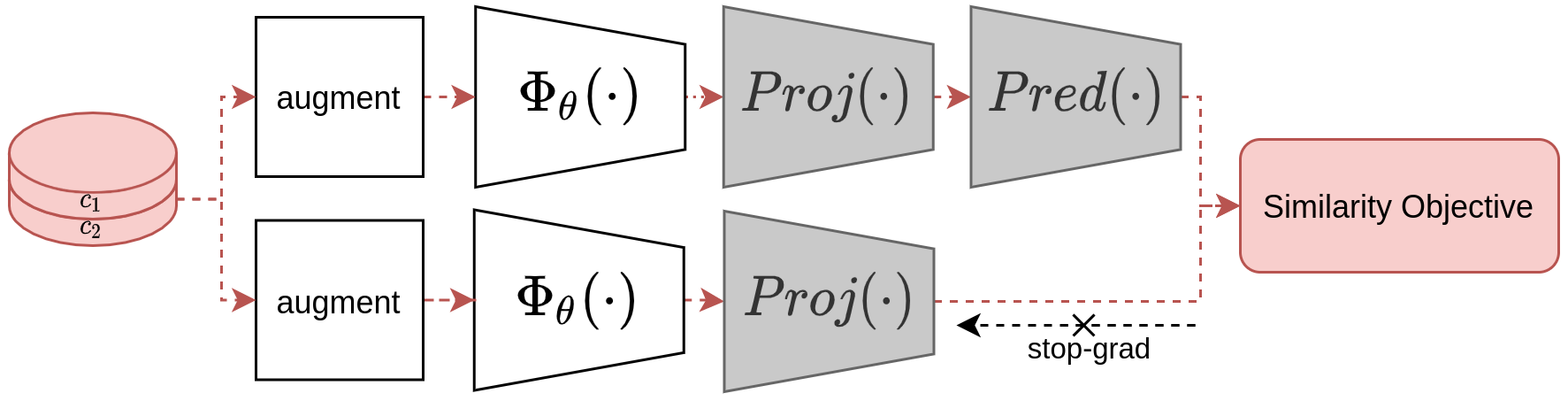} \\
                   \midrule
        \firstcol{Semi-Supervised} & \archsubfloat{\SemiSupervised}{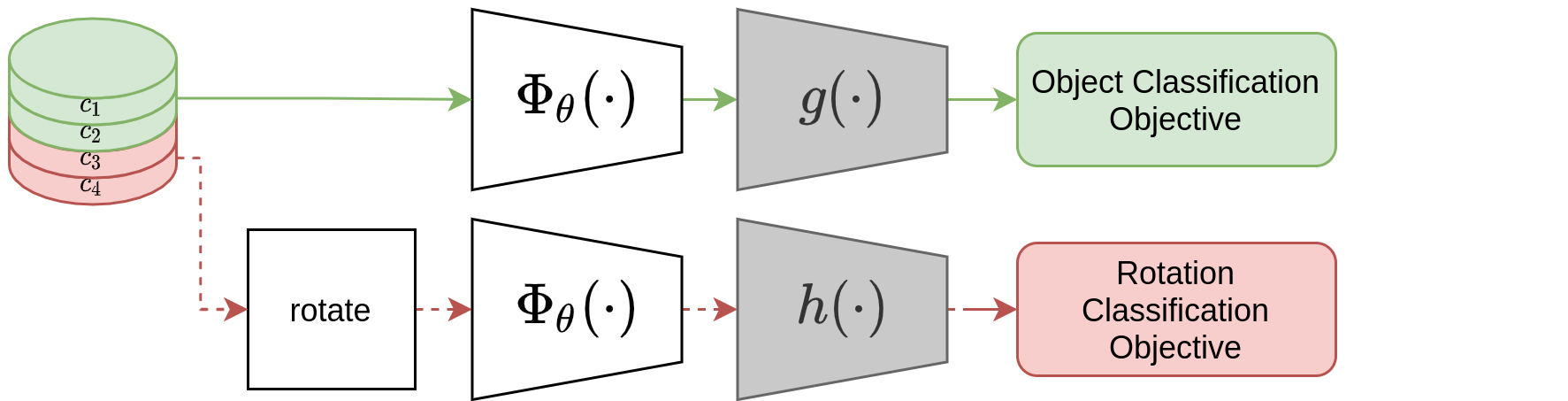} & \archsubfloat{\FixMatch}{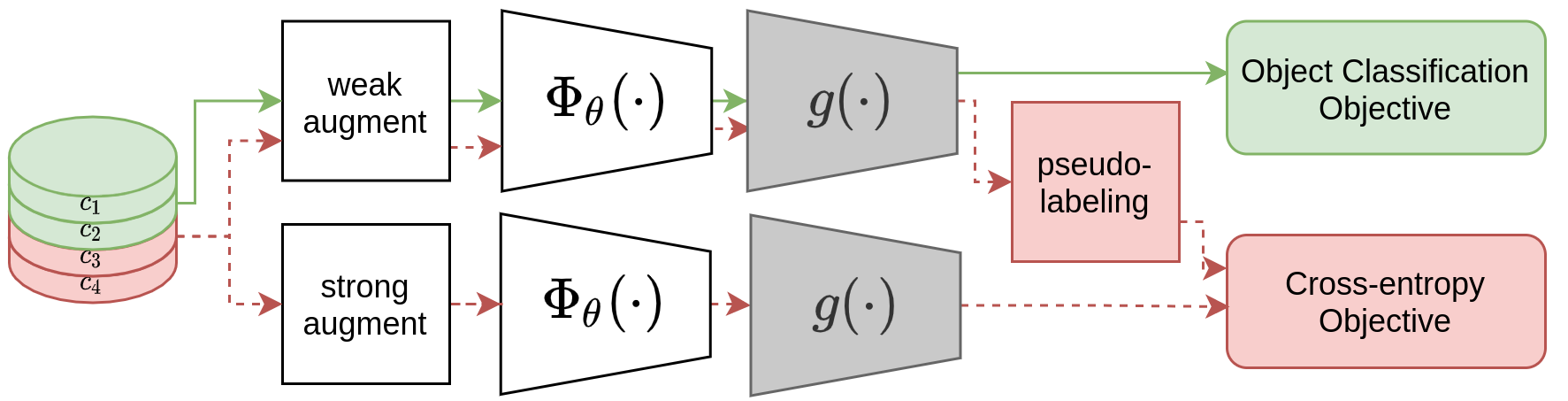} \\
        \bottomrule
    \end{tabular}
    \caption{\archcaption}
    \label{tab:algorithms}
\end{table*}
} { 
\begin{\algofigenv}[ht]
    \centering
    \archsubfloat{\PlainSupervised}{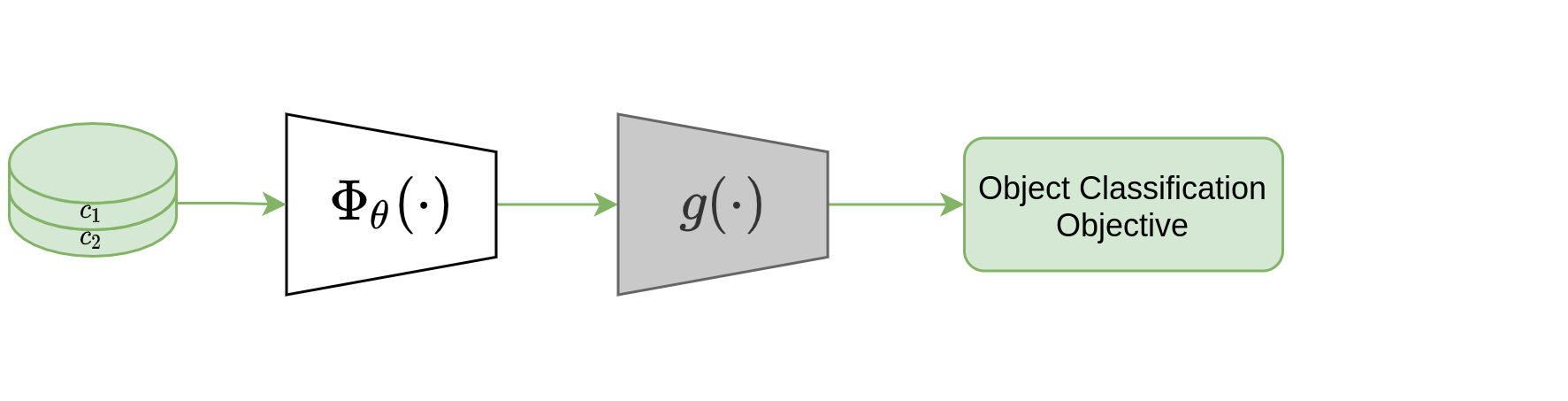}
    \archsubfloat{\TripletLoss}{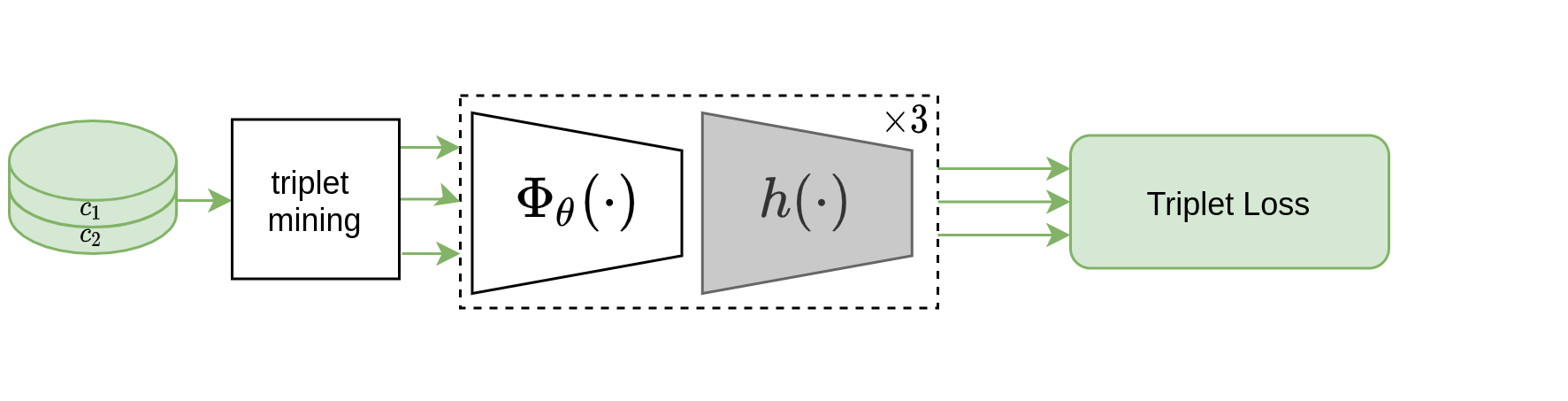}
    \\
    \archsubfloat{\ClassificationWithTripletLoss}{cwt}
    \archsubfloat{\SupContrast}{supcon}
    \\
    \archsubfloat{\UnsupervisedRotation}{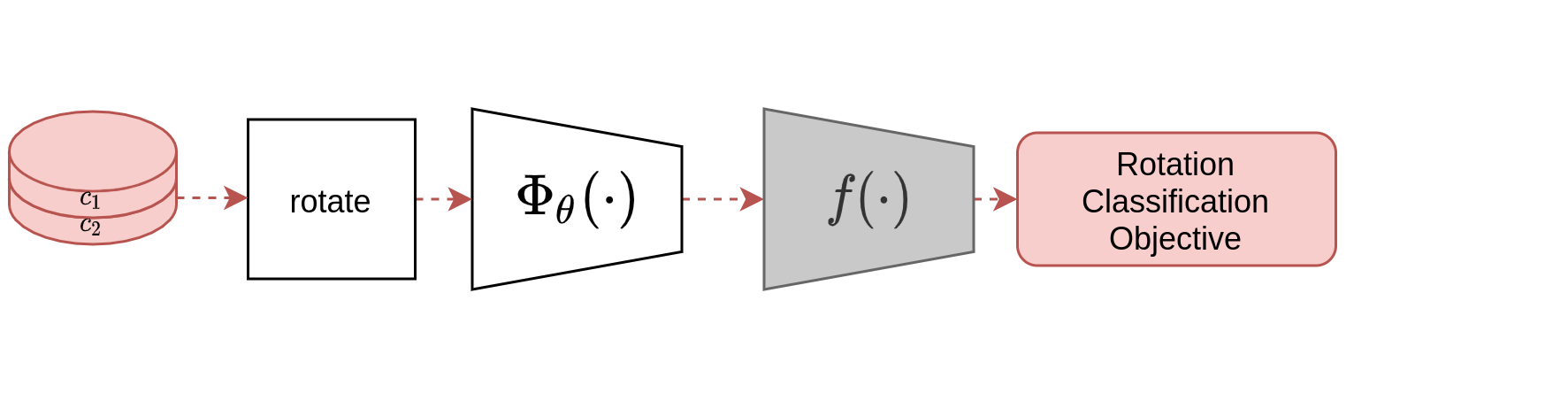}
    \archsubfloat{\SimSiam}{simsiam}
    \\
    \archsubfloat{\SemiSupervised}{s4ls}
    \archsubfloat{\FixMatch}{fixmatch}
    \caption{\archcaption}
    \label{fig:algorithms}
\end{\algofigenv}
}

\newcommand{\archfig}[2]{
\iftoggle{archall}{
\begin{figure}[h]
    \centering
    \includegraphics[width=\linewidth]{images/arch/#1}
    \caption{{#2}}
    \label{fig:arch-{#1}}
\end{figure}
}{}
}

In this section we present the algorithms we have implemented and tested in our experiments.
We experimented with three training schemes.
We took four supervised algorithms,
two unsupervised algorithms
and two semi-supervised algorithms.
The architecture of each algorithm are illustrated in~\tabref{tab:algorithms}.
In the following we briefly introduce each algorithm.

\subsection{Supervised Algorithms}

\paragraph{\PlainSupervisedLong}
\archfig{ps}{\ph{Plain Supervised}}
This model optimizes the cross-entropy loss on \BC only:
\begin{equation}
\cL_V = \sum_{(I,y) \in \xT} \ell(g(\net{I}), y).
\end{equation}

\paragraph{\TripletLossLong}
\archfig{t}{\ph{Triplet Loss}}
This model optimizes the triplet loss~\cite{schroff2015} on \BC only:
\begin{equation}
    \cL_T = 
    \sum_{(I_a, I_p, I_n) \in \cT}
    [
    d_{ap} - d_{an} + \alpha
    ]_{+},
\end{equation}
where
$\cT$ is the set of all triplets mined from $\xT$;
$\alpha$ is a margin hyperparameter,
$d_{ap}=\|h(\net{I_a}) - h(\net{I_p})\|_2^2$,
$d_{an}=\|h(\net{I_a}) - h(\net{I_n})\|_2^2$.
In practice we optimize $\cL_T$ with SGD and mine \textit{semi-hard} triplets within a batch.

\paragraph{\ClassificationWithTripletLossLong}
\archfig{cwt}{\ph{Classification With Triplet Loss}}
This model jointly optimizes the cross-entropy loss and the triplet loss:
\begin{equation}
    \cL_{CwT}= \cL_V + \cL_T.
\end{equation}
The first head $g$ classifies into the \BC while the additional head $h$ reduces the dimension
and is used to compute the triplet loss.

\paragraph{Supervised Contrastive: \SupContrast}
\archfig{supcon}{\ph{Supervised Contrastive Loss}}
This model~\cite{khosla2021} is based on a siamese architecture.
It extends self-supervised contrastive methods~\cite{he2019,chen2020,grill2020a}
to the fully supervised-setting: in addition to construct positive pairs with
random augmentations of the same image, it also uses the class labels to construct
positive pairs.
Similar to the triplet loss the contrastive loss enforces positive pairs to attract 
themselves and negative pairs to repel themselves.
We refer the reader to the paper~\cite{khosla2021} for details.

\subsection{Unsupervised Algorithms}

\paragraph{Predicting Image Rotations: RotNet}
\archfig{rot}{\ph{Predicting Image Rotations}}
This model~\cite{gidaris2018} is an instance of a self-supervised algorithm~\cite{jing2019}.
It rotates the input image into four possible rotations,
and the head $g$ classifies into the four possible rotations.
The model is trained to correctly classify the rotation applied to the input.
It is a $4$-class classification problem optimized with a cross-entropy loss:
\begin{equation}
    \cL_R = \sum_{(I,y)\in \xT} \sum_{r\in \cR}\ell(g(\net{I^r}),r),
\end{equation}
where $\cR=\{\ang{0},\ang{90},\ang{180},\ang{270}\}$ is the set of all possible rotations
and $I^r$ refers to the corresponding rotated input.

\paragraph{SimSiam}
\archfig{simsiam}{\ph{SimSiam Algorithm (Unsupervised)}}
This model~\cite{chen2021c} is another instance of a self-supervised algorithm~\cite{jing2019}.
It is based on a siamese architecture and is trained 
to maximize the similarity between two randomly augmented views of an input image.
Despite being simpler than similar methods
\cite{he2019,chen2020,grill2020a}
it shows competitive results for unsupervised visual representation learning~\cite{chen2021c}.
We trained our own models on \BC only.
We refer the reader to the paper~\cite{chen2021c} for details.

\subsection{Semi-Supervised Algorithms}
\paragraph{FixMatch}
\archfig{fixmatch}{\ph{FixMatch Algorithm (Semi-Supervised)}}
This semi-supervised model~\cite{sohn2020a} feeds an unlabeled image to
a weak augmentation and a strong augmentation:
the weakly augmented input is used for pseudo-labeling and 
the strongly augmented input is used for computing a cross-entropy loss against the pseudo-label.
Note that the head $g$ classifies into the \BC.
The pseudo-labels are among the \BC and
the unlabeled images that are pseudo-labeled are fed to $g$ that classifies them into the \BC.
The pseudo-labeling module is a threshold applied on the probability of the most confident class:
only those images that are classified most confidently are passed to the next cross-entropy loss.
\FixMatch has two main hyperparameters: 
$\mu$ the number of unlabeled images per labeled images and 
$\tau$ the confidence threshold for the pseudo-labeling.
We refer the reader to~\cite{sohn2020a} for details.

\paragraph{\SemiSupervisedLong: \SemiSupervised}
\archfig{s4ls}{\ph{Semi-Supervised S4L simplified}}
We propose to train a network in a semi-supervised fashion 
by adding an auxiliary self-supervised loss for unsupervised novel-class samples.
We append a second head to the network $h$ tasked to predict the rotation applied to the input image as in the RotNet model~\cite{gidaris2018}.
This model optimizes the loss:

\begin{equation}
    \cL_{CwR} = \sum_{(I, y) \in \xT} 
    \ell(g(\net{I}), y) + 
    \sum_{r \in \cR}\sum_{I \in \imT} 
    \ell(h(\net{I^r}), r).
\end{equation}

The choice for the auxiliary self-supervised objective was motivated by a series of works
that successfully applied it to improve few-shot classification~\cite{gidaris2019, su2019, chen2019a},
robustness~\cite{hendrycks2019},
pre-training for novel class discovery~\cite{han2021a},
image generation~\cite{chen2019b,lucic2019},
semi-supervised learning~\cite{zhai2019}.

This is actually a simplified version of the method proposed in~\cite{zhai2019}, where the consistency regularization and pseudo-labeling components of the training scheme are removed.

\section{Experimental Results} \label{sec:results}
\subsection{Datasets}
We experiment on \Cif, \CIF~\cite{krizhevsky2009} and a subset of \IN~\cite{russakovsky2015a}.
For each dataset, we split the label space randomly and semantically.
The splits are done once and for all.

\paragraph{\Cif}
For the \RS, we keep $5$ classes at random for \BC
and the remaining $5$ classes for \NC.
We got:
\begin{align*}
\cY_b & = \{
\vehicle{automobile},
\animal{bird},
\animal{deer},
\animal{frog},
\vehicle{ship}
\},\\
\cY_n & = \{
\vehicle{airplane},
\animal{cat},
\animal{dog},
\animal{horse},
\vehicle{truck}
\}.
\end{align*}
Note that $2$ vehicles and $3$ animals ended in \BC, 
and also $2$ vehicles and $3$ animals ended in \NC.
For the \SS, we also partitioned the label space evenly into
$5$ \BC and $5$ \NC; we chose the following partitioning:
\begin{align*}
\cY_b & = \{
\vehicle{airplane},
\vehicle{automobile},
\animal{bird},
\vehicle{ship},
\vehicle{truck}
\}, \\
\cY_n & = \{
\animal{cat},
\animal{dog},
\animal{deer},
\animal{frog},
\animal{horse}
\}.
\end{align*}
Note that $4$ vehicles and $1$ animal are in the \BC,
while the \NC consist only of $5$ animals and no vehicle.

\paragraph{\CIF}
\CIF contains $100$ classes that can be grouped into $20$ 
super-classes.
For the \RS, we keep $50$ classes at random for \BC
and the remaining $50$ classes for \NC.
Note that for the \RS, the super-classes are spread
randomly among the \B and the \N classes.
For example, the \textit{people} super-class has
$3$ classes in base classes (baby, girl, man) and
$2$ classes in novel classes (boy, woman).
For the \SS, we follow the splits of \FC100
introduced in~\cite{oreshkin2019}.
We set the \BC to the train classes
and the \NC to the val classes + test classes.
Note that for the \SS, each super-class is entirely either in
the \BC or the \NC.
For example, the \textit{people} super-class has all the $5$
classes in \NC.

\paragraph{\In} \label{sec:results-imagenet100}
We experiment on a subset of 100 classes of \IN~\cite{russakovsky2015a} which we call \In in the following.
We keep 16 of the high-level categories among the 34 devised by the authors of
\tiered~\cite{ren2018}
using the WordNet hierarchy:
8 categories descending from the \textit{artefact} synset 
(motor vehicle,
craft,
durables,
garment,
musical instrument,
game equipment,
furnishing,
tool),
8 categories descending from the \textit{animal} synset 
(ungulate,
primate,
feline,
working dog,
saurian,
aquatic bird,
insect,
aquatic vertebrate);
with 6-7 ImageNet classes per high-level category.
For the \RS we keep in \BC
3-4 classes for each high-level category at random in order to keep 50 classes, and
the \NC are the remaining 50 classes.
For the \SS, we keep the \textit{artefact} classes for the \BC
and the \textit{animal} classes for the \NC.
Unlike \tiered we keep all training images in the training, and validation images for testing; like \tiered we resize the images to 84x84 resolution.


\subsection{Implementation details}
All networks share the same ResNet18~\cite{he2016a} backbone.
On \Cif and \CIF, the input images are 32x32 in resolution,
we use a first convolutional layer with stride $1$,
not followed by a max pooling layer.
For \In, the images are stored on disk at 84x84 resolution and are
resized to 224x224 resolution before being fed to the network:
this time we used the standard ResNet18 architecture where the first 
layers reduce the spatial dimension.

We reimplemented from scratch \PlainSupervised, \TripletLoss, \UnsupervisedRotation, \ClassificationWithTripletLoss and \SemiSupervised.
For \PlainSupervised, \TripletLoss, \ClassificationWithTripletLoss and \SemiSupervised,
the networks are optimized with SGD with batch size $128$, 
$0.9$ momentum, 
$1e^{-4}$ weight-decay, 
and an initial learning rate of $0.1$ dropped by $10$ every $30$ epochs for a total of $100$ epochs.
For \UnsupervisedRotation, we followed the guidelines from the paper~\cite{gidaris2018}.
For \TripletLoss and \ClassificationWithTripletLoss,
in all cases we set the margin $\alpha$ to $0.1$
we mine \textit{semihard}~\cite{schroff2015} negative samples,
and we set the embedding dimension to $128$.

For the other three algorithms we used 
PyTorch implementations readily available 
(\SimSiam
\footnote{\href{https://github.com/facebookresearch/simsiam}{https://github.com/facebookresearch/simsiam}},
\SupContrast
\footnote{\href{https://github.com/HobbitLong/SupContrast}{https://github.com/HobbitLong/SupContrast}},
\FixMatch
\footnote{\href{https://github.com/kekmodel/FixMatch-pytorch}{https://github.com/kekmodel/FixMatch-pytorch}})
and made some modifications.
We adapted the codebase
by modifying the backbone network
to match the Resnet18 used for the other baseline algorithms.
For \SupContrast on \In we reduced the number of epochs to 200, decaying the initial learning rate by 10 at epochs 150, 170 and 190.
For \FixMatch on \In, we trained for $2^{19}$ iterations and used $\mu=5$ unlabeled images per labeled image.
We kept the recommended hyperparameters in every other case.
We would like to emphasize that we did not optimize the hyperparameters
for the best performance but instead use sensible defaults, following recommendations
of ResNet~\cite{he2016a} and original papers of each algorithm.

\subsection{Results and analysis}

\begin{table}[h]
	\centering
	\begin{tabular}{lrrrr}
		\toprule
		Algo & 
		\multicolumn{2}{c}{\Cif-Random} & 
		\multicolumn{2}{c}{\Cif-Semantic}
		\\
		\cmidrule(lr){2-3}  
		\cmidrule(lr){4-5}
  & Base & Novel & Base & Novel \\
\PlainSupervised              &65.147&26.380&68.469&19.758\\
\TripletLoss                  &71.479&20.512&74.890&16.760\\
\ClassificationWithTripletLoss&72.336&20.391&74.389&15.438\\
\SupContrast                  &\bestb{74.881}&28.954&\bestb{79.566}&22.726\\
\hline
\UnsupervisedRotation         &28.476&17.379&27.663&16.732\\
\SimSiam                      &19.340&19.048&20.518&17.298\\
\hline
\SemiSupervised               &72.816&37.832&77.680&\bestn{35.764}\\
\FixMatch                     &57.807&\bestn{39.238}&79.178&27.419\\

		\bottomrule
	\end{tabular}
    \caption{
    R-Precision on \B and \N classes of \Cif for
    a \RS (left columns) and
    a \SS (right columns).
    Top rows are supervised algorithms.
    Middle rows are unsupervised algorithms.
    Bottom rows are semi-supervised algorithms.
    Best R-Precision is marked in \textbf{bold}.}
	\label{tab:cifar10}
\end{table}

\begin{table}[h]
	\centering
	\begin{tabular}{lcccc}
		\toprule
		Algo & 
		\multicolumn{2}{c}{\CIF-Random} & 
		\multicolumn{2}{c}{\CIF-Semantic}
		\\
		\cmidrule(lr){2-3}  
		\cmidrule(lr){4-5}
 & Base & Novel & Base & Novel \\
\PlainSupervised              &33.339&15.351&39.778&7.891\\
\TripletLoss                  &33.895&10.015&43.745&3.846\\
\ClassificationWithTripletLoss&\bestb{48.656}&10.180&\bestb{57.021}&4.131\\
\SupContrast                  &42.981&14.094&49.000&6.678\\
\hline
\UnsupervisedRotation         &5.293 &3.906 &6.149 &2.392\\
\SimSiam                      &5.287 &4.521 &6.085 &3.863\\
\hline
\SemiSupervised               &34.257&17.525&39.962&\bestn{9.886}\\
\FixMatch                     &38.771&\bestn{22.896}&47.905&9.444\\

		\bottomrule
	\end{tabular}
    \caption{
    R-Precision on \B and \N classes of \CIF for
    a \RS (left columns) and
    a \SS (right columns).
    Top rows are supervised algorithms.
    Middle rows are unsupervised algorithms.
    Bottom rows are semi-supervised algorithms.
    Best R-Precision is marked in \textbf{bold}.}
	\label{tab:cifar100}
\end{table}

\begin{table}[h]
	\centering
	\begin{tabular}{lrrrr}
		\toprule
		Algo & 
		\multicolumn{2}{c}{\In-Random} & 
		\multicolumn{2}{c}{\In-Semantic}
		\\
		\cmidrule(lr){2-3}  
		\cmidrule(lr){4-5}
 & Base & Novel & Base & Novel \\

\PlainSupervised              &35.202&18.351&35.388&9.931 \\
\TripletLoss                  &36.680&12.018&27.138&3.905 \\
\ClassificationWithTripletLoss&\bestb{48.705}&14.090&\bestb{48.322}&4.813 \\
\SupContrast                  &33.495&15.241&34.667&7.738 \\
\hline
\SimSiam                      &7.984 &7.867 &8.359 &6.306 \\
\UnsupervisedRotation         &6.789 &7.194 &8.469 &5.194 \\
\hline
\SemiSupervised               &37.287&\bestn{20.403}&36.373&\bestn{13.238}\\
\FixMatch                     &38.090&19.322& 35.230      &  11.050    \\

		\bottomrule
	\end{tabular}
    \caption{
    R-Precision on \B and \N classes of \In for
    a \RS (left columns) and
    a \SS (right columns).
    Top rows are supervised algorithms.
    Middle rows are unsupervised algorithms.
    Bottom rows are semi-supervised algorithms.
    Best R-Precision is marked in \textbf{bold}.}
	\label{tab:imagenet100}
\end{table}

Retrieval performance is reported in
~\tabref{tab:cifar10} (\Cif),
~\tabref{tab:cifar100} (\CIF) and
~\tabref{tab:imagenet100} (\In).
Read horizontally we can compare the performance between \B and \N classes. 
In any case the performance is degraded on \NC, even more so for the \SS.
Read vertically we can compare the performance between the algorithms.

\paragraph{Supervised algorithms}
The metric learning algorithms (\TripletLoss and \ClassificationWithTripletLoss) exhibit a better fit of the \BC
to the detriment of a poorer performance on the \NC classes compared to \PlainSupervised.
The benefit of \TripletLoss over \PlainSupervised on \BC is most noteworthy on \Cif and slight on \CIF and \In,
indicating that the simpler and faster \PlainSupervised algorithm is a better choice on more fine-grained dataset.
The contrastive model \SupContrast performs the best among the supervised baselines on \BC;
on \NC it also performs the best on \Cif but not on \CIF and \In.
The contrastive model \SupContrast performs the best on \B and \N classes on \Cif but this does not hold on \CIF and \In.

\paragraph{Unsupervised algorithms}
\UnsupervisedRotation and \SimSiam do not show interesting results 
even on the \BC there were trained on;
and we do not draw a consistent conclusion as for which algorithm is best
in general.
However we note that for the semantic split in some cases \UnsupervisedRotation and \SimSiam are better than \TripletLoss and \ClassificationWithTripletLoss on novel classes.
For example on \In-Semantic it is actually better to use an unsupervised method than to use a metric learning method.

\paragraph{Semi-Supervised algorithms}
The semi-supervised models mitigate the 
performance difference between \B and \N classes
and surpass the supervised baselines with a greater margin
on the \SS than on the \RS.
We note that \SemiSupervised performs consistently better
than \PlainSupervised on all datasets on both \B and \N classes.

As for \FixMatch, it is consistently better than \PlainSupervised on \NC.
It compares more favorably to \SemiSupervised on \NC for the \RS than for the \SS.
We argue that it is due to the fact that during training it learns to classify \N-class images into the set of \BC and, because for the \SS the \NC are less visually alike to the \BC than for the \RS, it less efficiently leverages similar visual patterns between \B and \N classes.
This comparison between \SemiSupervised and \FixMatch for the \SS is an evidence that semi-supervised methods such as \FixMatch that rely on the fact that labeled and unlabeled images come from the same set of classes are less efficient in the setting when an entire set of images from a \N class come into play.



\subsection{T-SNE embedding visualization.}
\def\tsnewidth{0.20\textwidth}

\newcommand{\tsnesubfloat}[4]{
    \raisebox{-.5\height}{\includegraphics[width=\tsnewidth]{images/tsne/color2/#3/#4/#2}}
}

\renewcommand{\firstcol}[1]{
    \textbf{#1}
}

\newcommand{\captiontsne}{
T-SNE visualizations of the embeddings of 
\B and \N classes for \Cif 
for the \RS (left columns) 
and for the \SS (right columns).
The Semi-Supervised algorithms (bottom rows)
better separate the \NC compared to
the Supervised algorithms (top rows). 
The colors encode class membership. Best viewed in color.
}
\begin{table*}[ht]
    \centering
    \begin{tabular}{lcccc}
        \toprule
        &
		\multicolumn{2}{c}{\Cif-Random} & 
		\multicolumn{2}{c}{\Cif-Semantic}
		\\
		\cmidrule(lr){2-3}  
		\cmidrule(lr){4-5}
        & Base & Novel & Base & Novel
         \\
        \firstcol{\PlainSupervised} &
        \tsnesubfloat{\PlainSupervised}{ps}{random}{base} &
        \tsnesubfloat{\PlainSupervised}{ps}{random}{novel} &
        \tsnesubfloat{\PlainSupervised}{ps}{semantic}{base} &
        \tsnesubfloat{\PlainSupervised}{ps}{semantic}{novel} 
    \\
        \firstcol{\ClassificationWithTripletLoss} & 
        \tsnesubfloat{\ClassificationWithTripletLoss}{cwt}{random}{base} &
        \tsnesubfloat{\ClassificationWithTripletLoss}{cwt}{random}{novel} &
        \tsnesubfloat{\ClassificationWithTripletLoss}{cwt}{semantic}{base} &
    \tsnesubfloat{\ClassificationWithTripletLoss}{cwt}{semantic}{novel}
    \\
    \midrule
        \firstcol{\SemiSupervised} & 
        \tsnesubfloat{\SemiSupervised}{s4ls}{random}{base} &
        \tsnesubfloat{\SemiSupervised}{s4ls}{random}{novel} &
        \tsnesubfloat{\SemiSupervised}{s4ls}{semantic}{base} &
        \tsnesubfloat{\SemiSupervised}{s4ls}{semantic}{novel}
    \\
        \firstcol{\FixMatch} &
        \tsnesubfloat{\FixMatch}{fixmatch}{random}{base} &
        \tsnesubfloat{\FixMatch}{fixmatch}{random}{novel} &
        \tsnesubfloat{\FixMatch}{fixmatch}{semantic}{base} &
        \tsnesubfloat{\FixMatch}{fixmatch}{semantic}{novel}
        \\
        \bottomrule
    \end{tabular}
    \caption{\captiontsne}
    \label{tab:tsne}
\end{table*}

We show in~\tabref{tab:tsne} some T-SNE visualization of the embeddings of \Cif for four algorithms.
The visualizations support the conclusions from the previous section.
In particular, we can see that only the semi-supervised methods succeed in structuring
the \NC in the case of the \SS
(with CwRot providing a better class separation than \FixMatch).

\section{Conclusion}
We presented a method to evaluate novel class retrieval.
We argued that existing benchmarks for semi-supervised representation learning algorithms lack a setting where unlabeled data are from novel classes.

We experimented with a variety of representation learning algorithms
and showed evidence that semi-supervised learning algorithms mitigate
the performance drop on \NC.
Yet there is still much room for improvement for the \ncr
task. 
Semi-supervised algorithms still fall short in retrieving images of \NC 
in the \RS setting (low-degree of novelty) and even more so in the \SS setting (higher degree of novelty).

\appendix

{\small
\bibliographystyle{ieee_fullname}
\bibliography{main}
}

\end{document}